\documentclass[runningheads]{llncs}
\usepackage[T1]{fontenc}
\usepackage{graphicx,verbatim}
\usepackage{amsmath,amssymb}
\usepackage{booktabs}
\usepackage[table]{xcolor}
\usepackage{multirow}
\usepackage{algorithm}
\usepackage{algorithmic}
\usepackage{url}
\usepackage{hyperref}
\usepackage{makecell}
\usepackage{placeins}

\begin{document}
\title{LaSeD: Label-Semantic Self-Distillation for Visual-Only Surgical Phase Recognition}
\titlerunning{LaSeD for Surgical Phase Recognition}

%

\author{
Ye Tao\inst{1}
\and Claudia Scherl\inst{2,3}
\and Sara Monji-Azad\inst{1}\thanks{Corresponding author: sara.monjiazad@medma.uni-heidelberg.de}
}
\authorrunning{Y. Tao et al.}

\institute{
Mannheim Institute for Intelligent Systems in Medicine (MIISM), Medical Faculty Mannheim, Heidelberg University, Mannheim, Germany
\and Department of Otorhinolaryngology, Head and Neck Surgery, University Medical Center Mannheim, Medical Faculty Mannheim, Heidelberg University, Mannheim, Germany
\and Department of Otolaryngology, Head and Neck Surgery, Campus Klinikum Bielefeld Mitte, University Hospital OWL of Bielefeld University, Bielefeld, Germany
}
  
\maketitle              
\begin{abstract}
Surgical phase recognition maps each video frame to a clinically meaningful workflow phase, supporting context-aware assistance, documentation, and postoperative analysis. Most methods treat phase annotations only as class IDs, whereas recent surgical vision-language models often require additional video--text data, captions, or instruction tuning. We propose \emph{LaSeD}, a label-semantic self-distillation framework that uses phase names as privileged training-time context while retaining visual-only deployment without a ground-truth phase-name hint. LaSeD initializes a frozen teacher and a student from the same pretrained VLM checkpoint. The teacher receives the frame, a fixed task prompt, and the ground-truth phase-name hint; the student receives the same frame and prompt without the hint, and only its visual encoder is optimized. Training combines hard phase-token supervision with feature-level distillation from cached teacher representations. At inference, the teacher and hint are removed, and the student predicts one of the seven Cholec80 phases through constrained digit-token logits without an additional classifier head. On the Cholec80 evaluation split, LaSeD achieves 86.20\% accuracy, 77.75\% macro recall, 78.13\% macro precision, and 64.15\% macro Jaccard. Under the identical protocol, it improves a visual-only Qwen3-VL-4B baseline by 9.45, 9.42, 11.93, and 12.22 percentage points, respectively. Visual-only means that the image is the only sample-specific inference input, while all frames share the same fixed task prompt. These results suggest that phase names provide a useful low-cost signal for adapting VLMs to surgical workflow analysis. (The code will be published soon.)
\keywords{Surgical phase recognition \and Label-semantic distillation \and Vision-language models \and Self-distillation \and Surgical workflow analysis}
\end{abstract}
\section{Introduction and Related Work}

\textbf{Surgical phase recognition.} Surgical phase recognition (SPR) assigns each time step in a surgical video to a clinically meaningful workflow state and supports intraoperative assistance, documentation, training, and postoperative review~\cite{garrow2021machine,demir2023review,li2024survey}. Cholec80 is a widely used laparoscopic cholecystectomy benchmark with seven phases, from preparation to gallbladder retraction~\cite{twinanda2016endonet}. SPR remains challenging because different phases may share instruments, tissue appearance, and viewpoints, while the same phase varies across patients and surgeons. Temporal models reduce this ambiguity using video context~\cite{zhang2023surgical,pan2023surgical,yang2024surgformer,wu2025holistic}, but phase names are typically treated only as class indices. This leaves open whether ground-truth phase-name information can improve a deployable model that receives no sample-specific label text at inference.

\textbf{Language supervision for surgical understanding.} Vision-language models (VLMs) show that natural-language supervision can shape visual representations~\cite{radford2021clip,yu2022coca}. In surgery, this has motivated video-language pretraining and multimodal instruction tuning, including SurgVLP~\cite{yuan2025surgvlp}, procedure-aware pretraining~\cite{yuan2024procedure}, HecVL~\cite{yuan2024hecvl}, and SurgVLM~\cite{zeng2025surgvlm}. These methods demonstrate the value of language, but often require large-scale video--text data, generated captions, transcripts, or instruction data. LaSeD instead uses only the phase names already available in standard annotations.

\textbf{Knowledge distillation and label-guided teachers.} Knowledge distillation transfers information from a teacher to a student through soft predictions or intermediate features~\cite{hinton2015distilling}. In SPR, self-distillation has been used to transfer richer representations from a teacher to a student~\cite{mohammadi2024self}. Guan et al.~\cite{guan2024label} introduced a label-guided teacher that receives both video features and ground-truth class labels during training. Their teacher learns label-aware temporal representations through label--feature cross-attention and supervised contrastive learning, which are then distilled into a frame-only temporal student. These studies show that privileged label information can provide a richer training signal while preserving label-free inference.

These observations raise a question: can phase-name conditioning improve a visual-only VLM for surgical phase recognition? Here, \emph{visual-only} means that the image is the only sample-specific inference input, while all frames share a fixed prompt defining the candidate phases and output format. The deployed model requires neither an external teacher nor captions, reports, or ground-truth phase-name text. We address this question with \emph{LaSeD}, a label-semantic self-distillation framework. During training, a frozen teacher receives the frame, fixed prompt, and ground-truth phase-name hint, while a student initialized from the same checkpoint receives the frame and prompt without the hint. The student predicts the phase and aligns its representation with the phase-name-conditioned teacher. Our contributions are threefold: we extend privileged-label distillation to pretrained VLMs using human-readable phase names as natural-language context; introduce prompt-controlled self-distillation in which only the student visual encoder is optimized; and evaluate LaSeD on Cholec80 with a constrained digit-token interface, showing clear gains over a hard-label visual-only baseline under the same protocol. We do not claim overall SPR state of the art; our goal is to assess whether phase-name conditioning improves visual-only VLM adaptation.

\section{Method}

\textbf{Problem formulation.} We consider surgical phase recognition from sampled laparoscopic video frames. Let $\mathcal{V}=\{I_t\}_{t=1}^{T}$ denote a video sequence, where $I_t$ is a preprocessed RGB frame and $y_t\in\mathcal{Y}=\{1,\ldots,C\}$ its phase label. For Cholec80, $C=7$, and each class $c$ has a human-readable phase name $s_c$, such as \emph{Preparation} or \emph{Calot Triangle Dissection}. A fixed prompt lists the seven candidate phases and specifies a digit-only answer format. LaSeD provides only the correct phase name $s_{y_t}$ as privileged, sample-specific text to the teacher during training; no external captions, reports, transcripts, or additional annotations are used.

\textbf{Constrained phase-token interface.} Rather than adding a task-specific classifier, we formulate phase recognition as constrained next-token prediction. Each class $c$ is mapped to a candidate digit token $\tau_c$, using the same mapping for the visual-only baseline, teacher, and student. The implementation verifies that each candidate digit is a single vocabulary token and obtains its ID from the VLM tokenizer. Given a frame $I_t$ and complete chat prompt $q$, the autoregressive VLM $f_\theta$ produces vocabulary logits at each input position. We use the logits at the final prompt position $a(q)$, which predict the first answer token; the answer digit is not included in the input prefix. Retaining only the $C$ candidate-token logits gives
\begin{equation}
\mathbf{z}_t(q)=[z_{t,1}(q),\ldots,z_{t,C}(q)],
\qquad
z_{t,c}(q)=[f_\theta(I_t,q)]_{a(q),\tau_c},
\end{equation}
with the restricted distribution
\begin{equation}
p(y=c\mid I_t,q)=
\frac{\exp(z_{t,c}(q))}
{\sum_{j=1}^{C}\exp(z_{t,j}(q))}.
\end{equation}
The prediction is $\hat{y}_t=\arg\max_c p(y=c\mid I_t,q)$. This shared interface avoids randomly initialized classifier parameters.

\begin{figure}[t]
    \centering
    \includegraphics[width=0.9\textwidth]{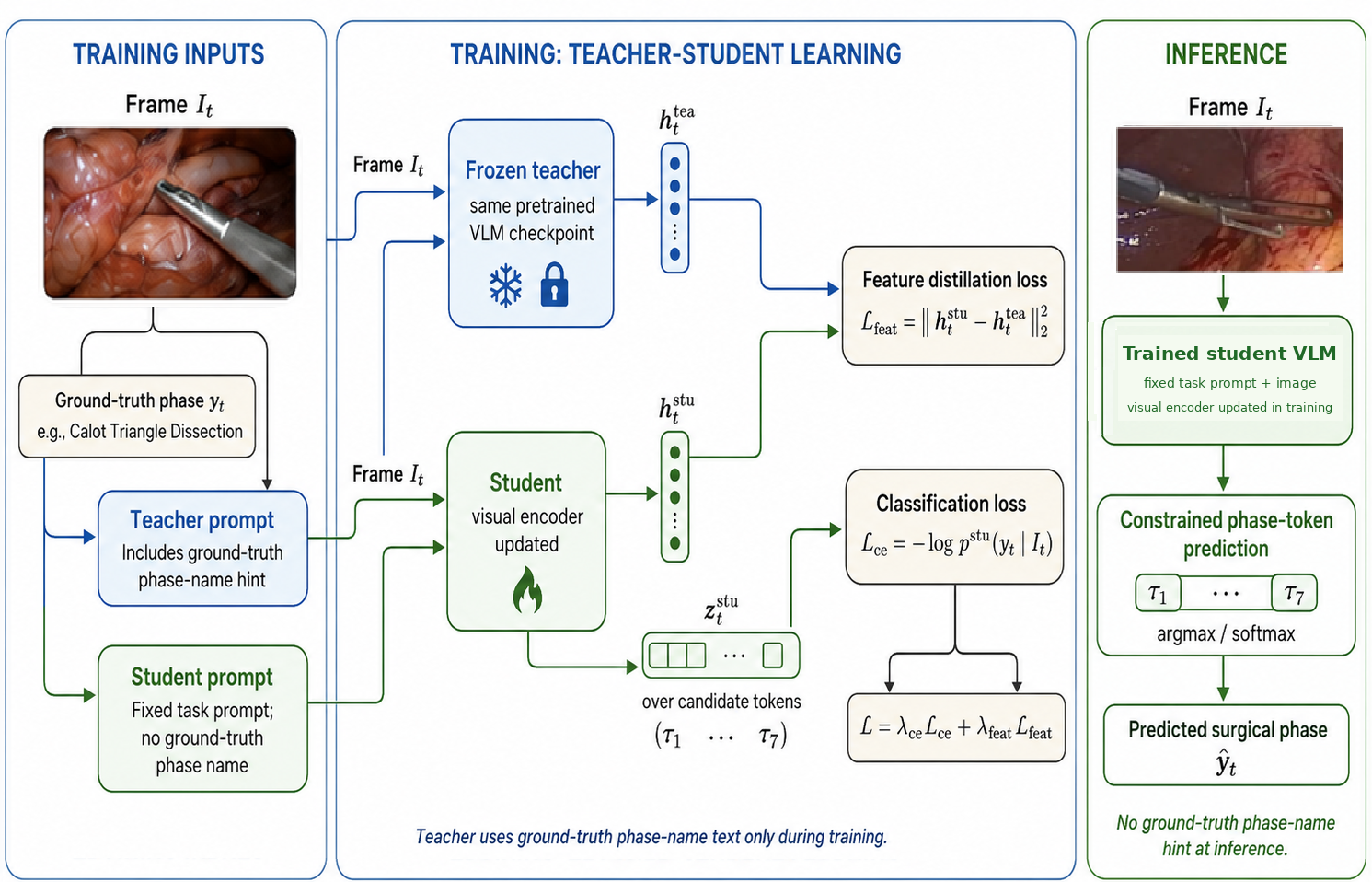}
    \caption{Overview of LaSeD. A frozen label-visible teacher provides representation targets to a visual-only student, which is trained with hard phase-token supervision and feature distillation. At inference, only the student and fixed prompt are retained.}
    \label{fig:method}
    \end{figure}

\textbf{Label-visible teacher and visual-only student.} Inspired by the privileged-label teacher--student principle of Guan et al.~\cite{guan2024label}, LaSeD adapts this idea to pretrained VLMs through prompt-controlled information visibility rather than training a dedicated label-aware temporal teacher. As shown in Figure~\ref{fig:method}, both branches are initialized from the same pretrained VLM checkpoint and use the same fixed system prompt, which lists the seven phases and requests one digit. The teacher $f_{\theta_T}$ is entirely frozen, whereas the student $f_{\theta_S}$ retains frozen language-model and output components and updates only its visual encoder.

For each training pair $(I_t,y_t)$, the student input contains the frame and fixed prompt $q^{\mathrm{stu}}$, with no ground-truth phase name or answer token. The teacher receives the same frame and task instruction together with the sample-specific statement ``This phase belongs to \{Label\},'' where \{Label\}=$s_{y_t}$, denoted by $q^{\mathrm{tea}}(s_{y_t})$. Thus,
\begin{equation}
x_t^{\mathrm{tea}}=\big(I_t,q^{\mathrm{tea}}(s_{y_t})\big),
\qquad
x_t^{\mathrm{stu}}=\big(I_t,q^{\mathrm{stu}}\big).
\end{equation}
The branches therefore differ only in access to the ground-truth phase-name hint, not in model size or architecture. Because the teacher receives the correct phase name, it serves only as a privileged-information branch that provides a label-conditioned representation target. The student learns to approximate this target from the frame and fixed prompt alone.

\textbf{Self-distillation objective.} Let $\mathbf{h}^{\mathrm{tea}}_{t,\ell},\mathbf{h}^{\mathrm{stu}}_{t,\ell}\in\mathbb{R}^{d}$ denote the language-model representations extracted at decoder layer $\ell$ from $x_t^{\mathrm{tea}}$ and $x_t^{\mathrm{stu}}$, respectively, where $d$ is the hidden dimension. Each representation is taken at its branch-specific final prompt position, whose logits predict the first answer token. Because the prompts differ in length, these positions are determined independently. Teacher representations are precomputed, cached, and detached during student optimization. For mini-batch $\mathcal{B}$ and distillation layers $\mathcal{S}$, the hard-label loss over the restricted phase-token distribution is

\begin{equation}
\mathcal{L}_{\mathrm{ce}}=
-\frac{1}{|\mathcal{B}|}
\sum_{t\in\mathcal{B}}
\log p_{\mathrm{stu}}(y=y_t\mid I_t,q^{\mathrm{stu}}),
\end{equation}
and the feature-distillation loss is
\begin{equation}
\mathcal{L}_{\mathrm{feat}}=
\frac{1}{|\mathcal{B}|\,|\mathcal{S}|\,d}
\sum_{t\in\mathcal{B}}
\sum_{\ell\in\mathcal{S}}
\left\|
\mathbf{h}^{\mathrm{stu}}_{t,\ell}
-
\operatorname{sg}\!\left(
\mathbf{h}^{\mathrm{tea}}_{t,\ell}
\right)
\right\|_2^2,
\end{equation}
where $\operatorname{sg}(\cdot)$ stops gradients through the teacher features. Teacher logits are not distilled. We use $\mathcal{S}=\{L\}$, where $L$ denotes the final language-model decoder layer. The complete objective is
\begin{equation}
\mathcal{L}=
\lambda_{\mathrm{ce}}\mathcal{L}_{\mathrm{ce}}
+
\lambda_{\mathrm{feat}}\mathcal{L}_{\mathrm{feat}},
\end{equation}
with $\lambda_{\mathrm{ce}}=\lambda_{\mathrm{feat}}=1.0$.

\textbf{Visual-only test-time prediction.} At inference, the teacher, ground-truth phase-name hint, and distillation loss are removed. The student receives $I_t$ with the same fixed prompt used for every frame, which defines the candidate phases and digit-only output format but contains no sample-specific label information. The predicted phase is selected by argmax over the seven candidate-token probabilities. Thus, the frame is the only sample-specific inference input, and LaSeD retains the same deployment interface as the visual-only baseline.

\section{Experiments and Results}

\textbf{Dataset and protocol.} We evaluate on Cholec80~\cite{twinanda2016endonet}, using the first 40 videos for training and the remaining 40 as an evaluation split. Sampling at 1~fps yields 86,304 training and 98,193 evaluation frames. During training, the evaluation split was monitored periodically, and the checkpoint with the highest evaluation accuracy was retained separately for each method. The absolute results may therefore be optimistic and should be viewed as exploratory; however, the comparison remains controlled because both methods use the same data division, evaluation schedule, and checkpoint-selection procedure. We denote the seven phases as P1--P7: Preparation, Calot Triangle Dissection, Clipping and Cutting, Gallbladder Dissection, Gallbladder Packaging, Cleaning and Coagulation, and Gallbladder Retraction. We report accuracy, macro recall, macro precision, and macro Jaccard, emphasizing macro metrics because of phase imbalance.

\textbf{Implementation details.} All experiments use Qwen3-VL-4B-Instruct~\cite{qwenteam2025qwen3vl}, with teacher, student, and baseline initialized from the same checkpoint. The fixed prompt maps the seven phase names to digits 1--7 and requests a one-digit answer. The student user input contains only the image, whereas the teacher additionally receives the ground-truth phase-name hint. All compared Qwen3-VL models use the same data division, sampling, candidate tokens, preprocessing, inference prompt, and checkpoint-selection rule. Models were evaluated every 500 optimization steps, with checkpoint selection based on evaluation accuracy. With eight GPUs, per-device batch size 1, and gradient accumulation over eight steps, the effective global batch size was 64. Images are resized to shortest edge 224 while preserving aspect ratio. Neither LoRA nor QLoRA is used; the language model and output components remain frozen, and only the student visual encoder is updated. The frozen teacher caches the final-decoder representation at the answer-prediction position and is then removed from memory. The baseline uses only hard phase-token supervision, whereas LaSeD adds feature distillation with $\lambda_{\mathrm{ce}}=\lambda_{\mathrm{feat}}=1.0$. We use AdamW with cosine scheduling, linear warmup, learning rate $1\times10^{-5}$, weight decay 0.01, two epochs, per-device batch size 1, and gradient accumulation over eight steps. Training took approximately 19.5 hours on 8 NVIDIA H20 GPUs; inference required approximately 3.5 hours on one NVIDIA RTX 5090, or 0.13 seconds per frame.

\textbf{Quantitative comparison.} Table~\ref{tab:main} contextualizes LaSeD against general-purpose and surgically adapted VLMs. Non-Qwen3-VL results are taken from the SurgVLM benchmark~\cite{zeng2025surgvlm}. Because these models differ in scale, adaptation strategy, output interface, and evaluation protocol, their results are contextual rather than strictly controlled. The primary evidence comes from the two Qwen3-VL-4B rows, which share the same model family, data division, preprocessing, digit-token interface, inference prompt, and checkpoint-selection rule. LaSeD achieves 86.20\% accuracy, 77.75\% macro recall, 78.13\% macro precision, and 64.15\% macro Jaccard, improving the visual-only baseline by 9.45, 9.42, 11.93, and 12.22 percentage points, respectively. The largest relative improvement is observed for macro Jaccard (+23.53\%).

\begin{table}[t]
\centering
\caption{Contextual VLM comparison on Cholec80. Values are percentages; MCQ denotes constrained candidate-token prediction and OV open-vocabulary evaluation. Non-Qwen3-VL results are taken from SurgVLM~\cite{zeng2025surgvlm} and are contextual because evaluation settings differ. The Qwen3-VL-4B baseline and LaSeD share the same model family, data division, preprocessing, output interface, prompt and checkpoint selection.}
\label{tab:main}
\scriptsize
\setlength{\tabcolsep}{3.2pt}
\begin{tabular}{p{5.2cm}lcccc}
\toprule
Model & Eval & Accuracy & Recall & Precision & Jaccard \\
\midrule
LLaVA-1.5-7B~\cite{liu2023visual} & MCQ & 23.46 & 14.22 & 11.53 & 6.31 \\
Phi-4-Multimodal-Instruct~\cite{abouelenin2025phi4multimodal} & MCQ & 22.45 & 14.83 & 12.16 & 5.55 \\
Mistral-Small-3.1-24B-Instruct-2503~\cite{mistral2025small31} & MCQ & 22.61 & 19.87 & 26.55 & 9.92 \\
InternVL3-8B~\cite{zhu2025internvl3} & MCQ & 23.88 & 15.21 & 15.19 & 6.67 \\
InternVL3-78B~\cite{zhu2025internvl3} & MCQ & 27.32 & 24.03 & 30.48 & 13.30 \\
MiniCPM-V-2.6~\cite{yao2024minicpmv} & MCQ & 15.20 & 12.84 & 13.79 & 5.82 \\
MiniCPM-O-2.6~\cite{openbmb2024minicpmo26} & MCQ & 17.75 & 19.68 & 29.24 & 9.22 \\
Gemma3-27B-it~\cite{gemmateam2025gemma3} & MCQ & 14.08 & 17.49 & 19.67 & 4.45 \\
Skywork-R1V-38B~\cite{peng2025skyworkr1v} & MCQ & 6.37 & 14.67 & 13.32 & 1.38 \\
Llama-4-Scout-17B-16E-Instruct~\cite{meta2025llama4} & MCQ & 35.77 & 14.38 & 11.34 & 5.33 \\
Qwen2.5-VL-7B-Instruct~\cite{qwenteam2025qwen25vl} & MCQ & 30.45 & 16.69 & 26.17 & 7.73 \\
Qwen2.5-VL-32B-Instruct~\cite{qwenteam2025qwen25vl} & MCQ & 37.23 & 23.05 & 29.63 & 13.30 \\
Qwen2.5-VL-72B-Instruct~\cite{qwenteam2025qwen25vl} & MCQ & 29.30 & 19.94 & 27.75 & 10.50 \\
GPT-4o-2024-0806~\cite{hurst2024gpt4o} & MCQ & 36.43 & 31.02 & 32.99 & 19.55 \\
Qwen 2.5 Max~\cite{qwenteam2025qwen25vl} & MCQ & 34.79 & 17.83 & 31.41 & 8.91 \\
Gemini 2.0 Flash~\cite{geminiteam2025gemini} & MCQ & 38.89 & 36.76 & 40.03 & 21.47 \\
SurgVLM-7B Full-tuning~\cite{zeng2025surgvlm} & OV & 70.30 & 61.90 & 59.76 & 43.86 \\
SurgVLM-32B Freeze-tuning~\cite{zeng2025surgvlm} & OV & 71.20 & 67.87 & 63.61 & 48.89 \\
SurgVLM-72B LoRA-tuning~\cite{zeng2025surgvlm} & MCQ & 69.66 & 68.10 & 68.70 & 50.18 \\
SurgVLM-72B LoRA-tuning~\cite{zeng2025surgvlm} & OV & 76.40 & 70.84 & 66.02 & 52.45 \\
Ours: Qwen3-VL-4B Visual-only baseline & MCQ & 76.75 & 68.33 & 66.20 & 51.93 \\
Ours: Qwen3-VL-4B LaSeD & MCQ & \textbf{86.20} & \textbf{77.75} & \textbf{78.13} & \textbf{64.15} \\
\midrule
\multicolumn{2}{l}{\emph{Absolute gain over the visual-only baseline}} & +9.45 & +9.42 & +11.93 & +12.22 \\
\multicolumn{2}{l}{\emph{Relative gain over the visual-only baseline}} & +12.31\% & +13.79\% & +18.02\% & +23.53\% \\
\bottomrule
\end{tabular}
\end{table}

\begin{table}[t]
\centering
\caption{Video-level performance over the 40 evaluation videos, reported as mean $\pm$ standard deviation across videos. Each model was trained once.}
\label{tab:video_variability}
\scriptsize
\setlength{\tabcolsep}{3.5pt}
\begin{tabular}{lcccc}
\toprule
Method & Accuracy & Recall & Precision & Jaccard \\
\midrule
Visual-only baseline & $76.91 \pm 10.04$ & $68.40 \pm 7.75$ & $65.87 \pm 8.02$ & $51.51 \pm 8.82$ \\
LaSeD & $\mathbf{86.55 \pm 5.39}$ & $\mathbf{78.18 \pm 7.26}$ & $\mathbf{76.82 \pm 7.52}$ & $\mathbf{63.92 \pm 8.37}$ \\
\bottomrule
\end{tabular}
\end{table}

\begin{figure}[htp]
    \centering
    \begin{minipage}[t]{0.46\textwidth}
        \centering
        \includegraphics[width=\linewidth]{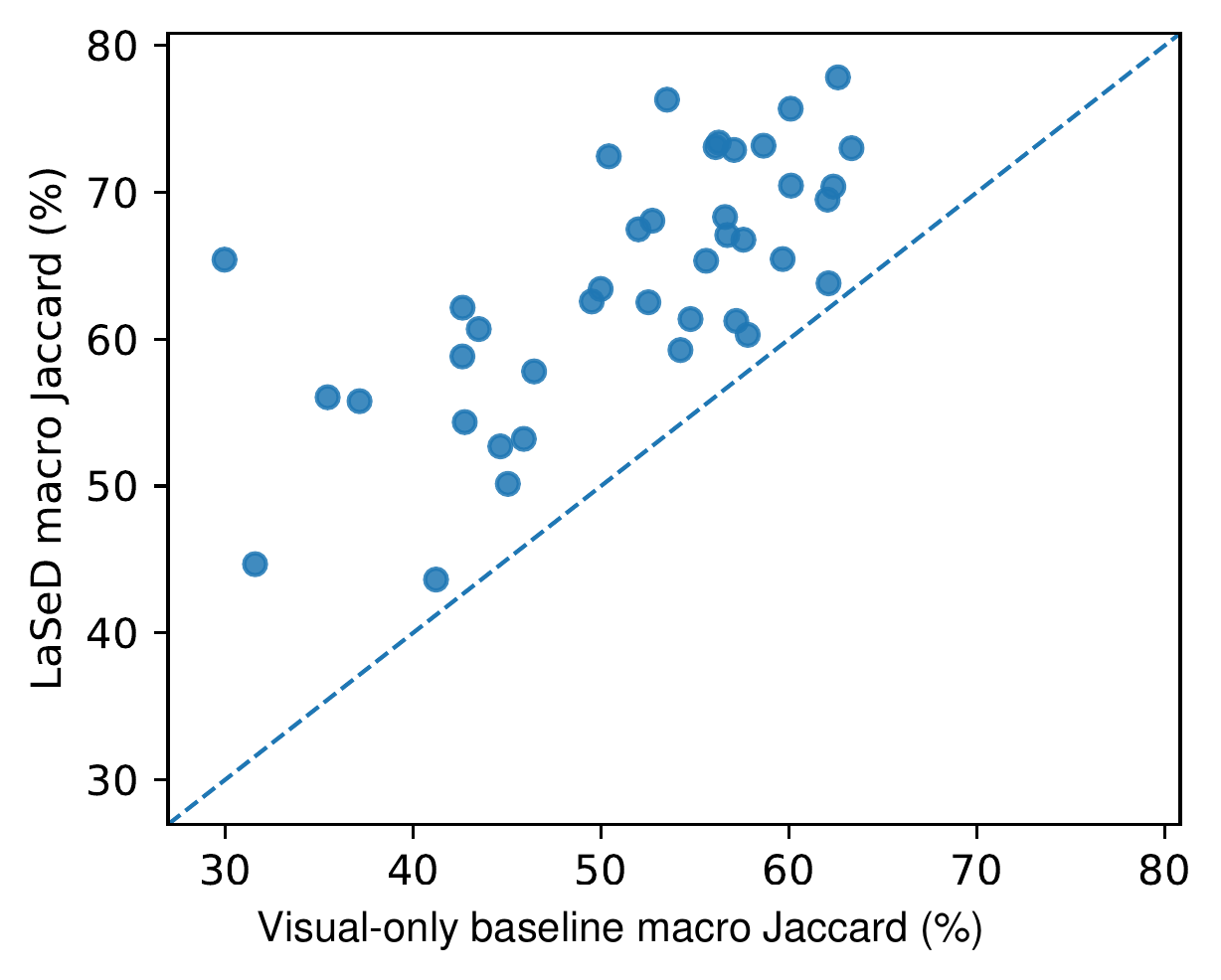}
        \small (a) Per-video macro Jaccard
    \end{minipage}\hfill
    \begin{minipage}[t]{0.52\textwidth}
        \centering
        \includegraphics[width=\linewidth]{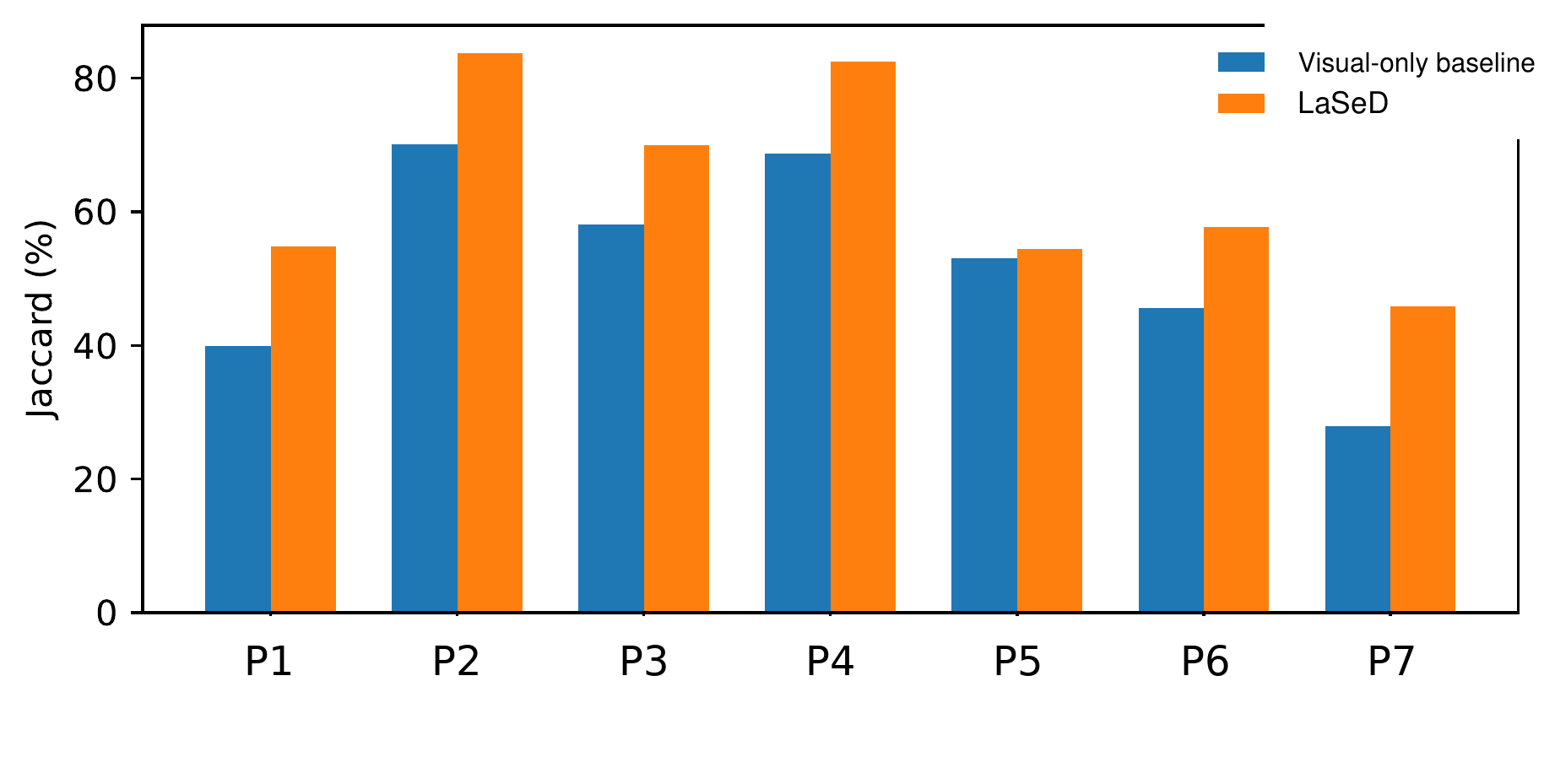}
        \small (b) Per-phase Jaccard
    \end{minipage}
    \caption{Additional analysis on the 40 evaluation videos: (a) per-video macro Jaccard and (b) per-phase Jaccard.}
    \label{fig:additional_eval}
\end{figure}

\begin{figure}[t]
    \centering
    \includegraphics[width=0.9\textwidth]{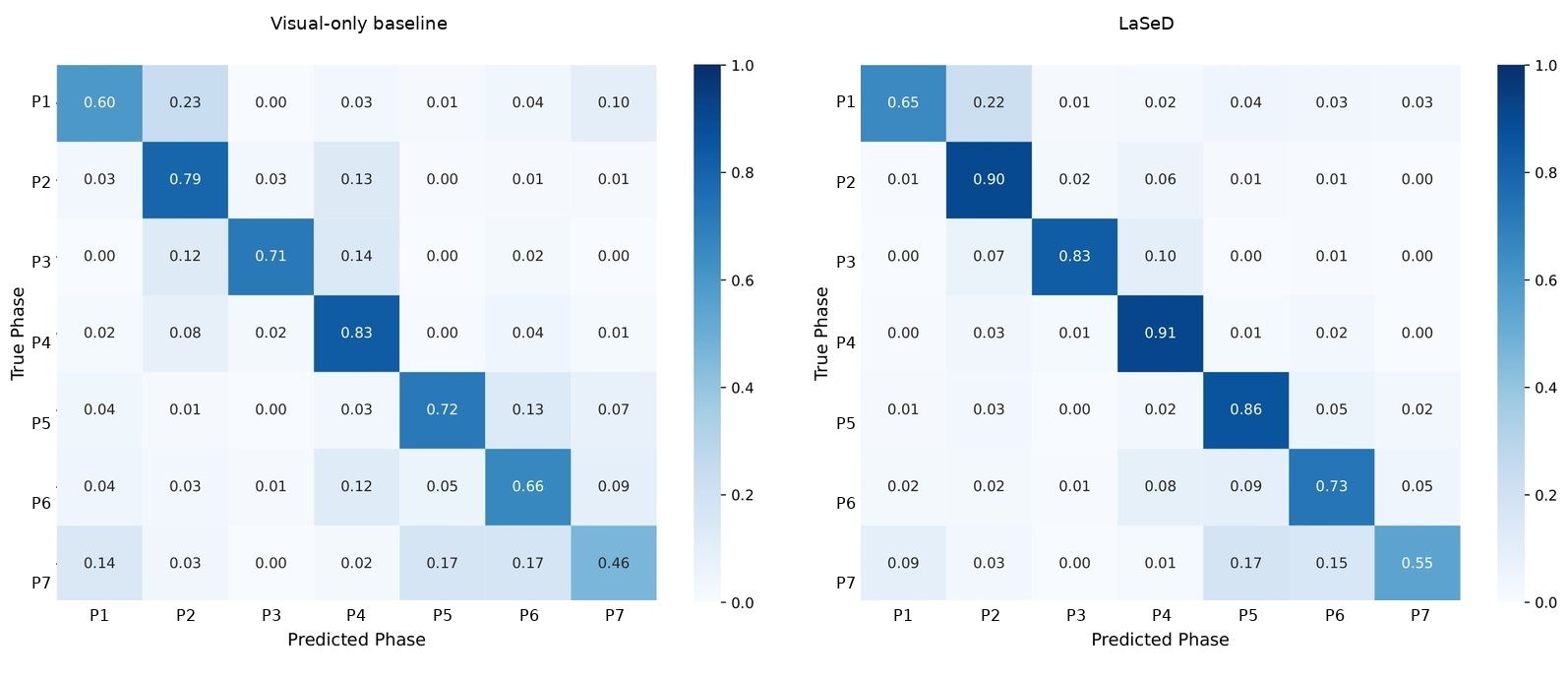}
    \caption{Normalized confusion matrices. LaSeD strengthens most diagonal entries, while visually similar late phases remain challenging.}
    \label{fig:confusion}
\end{figure}

\begin{figure}[t]
    \centering
    \includegraphics[width=0.8\textwidth]{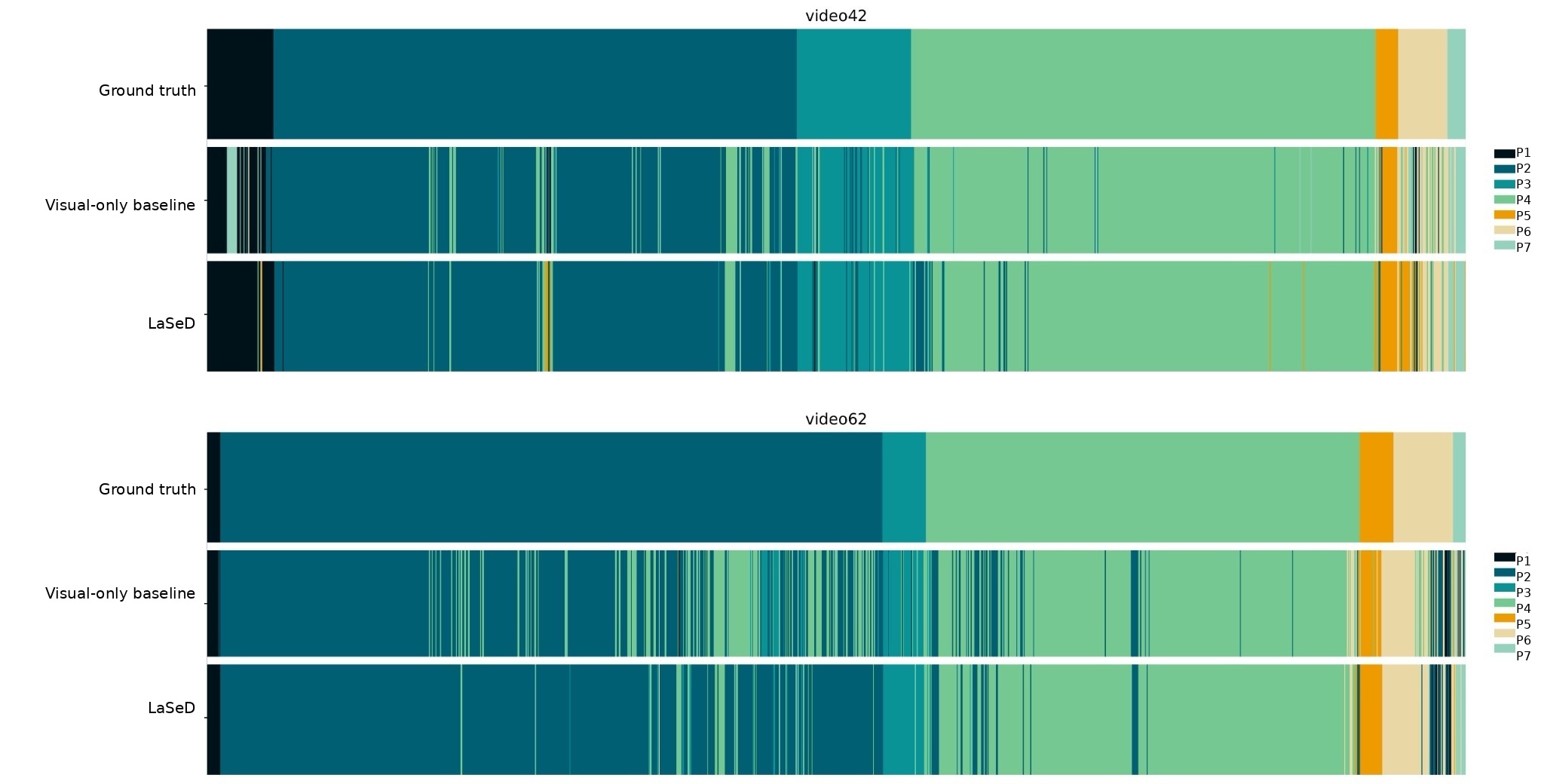}
    \caption{Representative phase predictions for two evaluation videos. Rows show ground truth, the visual-only baseline, and LaSeD.}
    \label{fig:phase_bars}
\end{figure}

\textbf{Video-level and phase-wise analysis.} Table~\ref{tab:video_variability} reports variability across the 40 evaluation videos. Per-video values differ slightly from the global frame-level metrics in Table~\ref{tab:main}. LaSeD improves the mean of every metric, reduces accuracy variability, and outperforms the baseline in accuracy and macro Jaccard on all 40 videos. These statistics characterize variation across videos, not training seeds. Figure~\ref{fig:additional_eval} shows that LaSeD also improves Jaccard for all seven phases, with the largest gains for P7, P1, P4, and P2. The smallest gain occurs for P5, where higher recall but lower precision indicates additional false positives for this short, visually similar late phase.

\textbf{Confusion patterns and temporal behavior.} Figure~\ref{fig:confusion} shows stronger diagonal responses for LaSeD, indicating improved phase separation. Remaining errors mainly involve the temporally adjacent phases P5--P7, which can appear similar in isolated frames. The phase bars in Figure~\ref{fig:phase_bars} show more coherent predictions than the baseline, although residual fluctuations suggest that explicit temporal modeling could further improve workflow consistency.

\section{Discussion and Conclusion}

\textbf{Why phase-name conditioning helps.} The controlled comparison shows that adding a phase-name-conditioned teacher target provides a richer training signal than hard phase-token supervision alone. The frozen teacher combines the surgical frame with the phase-name hint, producing a representation that reflects visual evidence and class-level context. By aligning with this representation, the student learns phase-discriminative visual features while operating without the hint at inference. This is especially useful when phases share instruments, anatomical appearance, or viewpoints, where hard class supervision may be insufficient. LaSeD therefore exploits semantic information already present in standard annotations without requiring captions, reports, text data, or a separately trained teacher.

\textbf{Consistency and practical implications.} LaSeD improves all four global metrics, with the largest relative gain in macro Jaccard, indicating better class-wise separation rather than improvement driven only by dominant phases. The gain is consistent across all 40 evaluation videos and all seven phases, showing that the benefit is broadly distributed. The phase-bar visualizations show more coherent predictions than the visual-only baseline, while the confusion matrices reveal stronger diagonal responses for most phases. Importantly, these improvements do not increase deployment-time input requirements: the teacher and ground-truth phase-name hint are removed, and inference uses only the trained student with the same fixed prompt as the baseline. The approach therefore adds training supervision without requiring a teacher model, external text, or sample-specific language at deployment.

\textbf{Remaining challenges and outlook.}
Most residual errors occur among the temporally adjacent late phases P5--P7, which can be difficult to distinguish from isolated frames. Combining LaSeD with temporal sequence modeling or lightweight smoothing may further improve workflow consistency. The present experiments establish the benefit of label-visible teacher conditioning with phase names, but do not isolate whether meaningful linguistic semantics outperform arbitrary class identifiers. Future studies should therefore compare phase-name hints with numeric-label, shuffled-name, and label-only teacher variants, as well as evaluate additional procedures, VLM families, and multiple training seeds. Because the evaluation split was used for checkpoint selection, future work should employ separate validation and untouched test sets.

\bibliographystyle{splncs04}
\bibliography{references}

@article{garrow2021machine,
  title={Machine Learning for Surgical Phase Recognition: A Systematic Review},
  author={Garrow, C. R. and Kowalewski, K. F. and Li, L. and Wagner, M. and Schmidt, M. W. and Engelhardt, S. and Hashimoto, D. A. and Kenngott, H. G. and Bodenstedt, S. and Speidel, S. and M{\"u}ller-Stich, B. P. and Nickel, F.},
  journal={Annals of Surgery},
  volume={273},
  number={4},
  pages={684--693},
  year={2021},
  doi={10.1097/SLA.0000000000004425}
}

@article{demir2023review,
  title={Deep Learning in Surgical Workflow Analysis: A Review of Phase and Step Recognition},
  author={Demir, Kubilay Can and Schieber, Hannah and Weise, Tobias and Roth, Daniel and Maier, Andreas and Yang, Seung Hee},
  journal={IEEE Journal of Biomedical and Health Informatics},
  year={2023},
  doi={10.1109/JBHI.2023.3311628}
}

@article{li2024survey,
  title={Deep Learning for Surgical Workflow Analysis: A Survey of Progresses, Limitations and Trends},
  author={Li, Y. and others},
  journal={Artificial Intelligence Review},
  year={2024},
  doi={10.1007/s10462-024-10929-6}
}

@inproceedings{twinanda2016endonet,
  title={EndoNet: A Deep Architecture for Recognition Tasks on Laparoscopic Videos},
  author={Twinanda, Andru P. and Shehata, Sherif and Mutter, Didier and Marescaux, Jacques and de Mathelin, Michel and Padoy, Nicolas},
  booktitle={Medical Image Computing and Computer-Assisted Intervention -- MICCAI 2016},
  pages={575--582},
  year={2016},
  publisher={Springer}
}

@article{zhang2023surgical,
  author  = {Zhang, Bokai and Goel, Bharti and Sarhan, Mohammad Hasan and Goel, Varun K. and Abukhalil, Rami and Kalesan, Bindu and Stottler, Nathan and Petculescu, Svetlana},
  title   = {Surgical Workflow Recognition with Temporal Convolution and Transformer for Action Segmentation},
  journal = {International Journal of Computer Assisted Radiology and Surgery},
  volume  = {18},
  number  = {4},
  pages   = {785--794},
  year    = {2023},
  doi     = {10.1007/s11548-022-02811-z}
}

@article{pan2023surgical,
  author  = {Pan, Xiaoying and Gao, Xuanrong and Wang, Hongyu and Zhang, Wuxia and Mu, Yuanzhen and He, Xianli},
  title   = {Temporal-Based Swin Transformer Network for Workflow Recognition of Surgical Video},
  journal = {International Journal of Computer Assisted Radiology and Surgery},
  volume  = {18},
  number  = {1},
  pages   = {139--147},
  year    = {2023},
  doi     = {10.1007/s11548-022-02785-y}
}

@inproceedings{yang2024surgformer,
  author    = {Yang, Shu and Luo, Luyang and Wang, Qiong and Chen, Hao},
  title     = {Surgformer: Surgical Transformer with Hierarchical Temporal Attention for Surgical Phase Recognition},
  booktitle = {Medical Image Computing and Computer Assisted Intervention -- MICCAI 2024},
  pages     = {606--616},
  year      = {2024},
  publisher = {Springer},
  doi       = {10.1007/978-3-031-72089-5_57}
}

@article{wu2025holistic,
  title={Holistic Surgical Phase Recognition with Hierarchical Input Dependent State Space Models},
  author={Wu, Haoyang and Wang, Tsun-Hsuan and Lechner, Mathias and Hasani, Ramin and Eckhoff, Jennifer A. and Pak, Paul and Meireles, Ozanan R. and Rosman, Guy and Ban, Yutong and Rus, Daniela},
  journal={arXiv preprint arXiv:2506.21330},
  year={2025}
}

@inproceedings{radford2021clip,
  title={Learning Transferable Visual Models From Natural Language Supervision},
  author={Radford, Alec and Kim, Jong Wook and Hallacy, Chris and Ramesh, Aditya and Goh, Gabriel and Agarwal, Sandhini and Sastry, Girish and Askell, Amanda and Mishkin, Pamela and Clark, Jack and others},
  booktitle={International Conference on Machine Learning},
  pages={8748--8763},
  year={2021},
  publisher={PMLR}
}

@article{yu2022coca,
  title={CoCa: Contrastive Captioners are Image-Text Foundation Models},
  author={Yu, Jiahui and Wang, Zirui and Vasudevan, Vijay and Yeung, Legg and Seyedhosseini, Mojtaba and Wu, Yonghui},
  journal={arXiv preprint arXiv:2205.01917},
  year={2022}
}

@article{yuan2025surgvlp,
  title={Learning Multi-Modal Representations by Watching Hundreds of Surgical Video Lectures},
  author={Yuan, Kun and Srivastav, Vinkle and Yu, Tong and Lavanchy, Joel L. and Marescaux, Jacques and Mascagni, Pietro and Navab, Nassir and Padoy, Nicolas},
  journal={Medical Image Analysis},
  year={2025},
  note={SurgVLP; arXiv:2307.15220}
}

@inproceedings{yuan2024procedure,
  title={Procedure-Aware Surgical Video-Language Pretraining with Hierarchical Knowledge Augmentation},
  author={Yuan, Kun and Srivastav, Vinkle and Navab, Nassir and Padoy, Nicolas},
  booktitle={Advances in Neural Information Processing Systems},
  year={2024}
}

@inproceedings{yuan2024hecvl,
  title={HecVL: Hierarchical Video-Language Pretraining for Zero-Shot Surgical Phase Recognition},
  author={Yuan, Kun and Srivastav, Vinkle and Navab, Nassir and Padoy, Nicolas},
  booktitle={Medical Image Computing and Computer-Assisted Intervention -- MICCAI 2024},
  pages={306--316},
  year={2024},
  publisher={Springer}
}

@article{zeng2025surgvlm,
  title={Surgvlm: A large vision-language model and systematic evaluation benchmark for surgical intelligence},
  author={Zeng, Zhitao and Zhuo, Zhu and Jia, Xiaojun and Zhang, Erli and Wu, Junde and Zhang, Jiaan and Wang, Yuxuan and Low, Chang Han and Jiang, Jian and Zheng, Zilong and others},
  journal={arXiv preprint arXiv:2506.02555},
  year={2025}
}

@inproceedings{hinton2015distilling,
  title={Distilling the Knowledge in a Neural Network},
  author={Hinton, Geoffrey and Vinyals, Oriol and Dean, Jeff},
  booktitle={NIPS Deep Learning and Representation Learning Workshop},
  year={2015}
}

@article{mohammadi2024self,
  title={Self-Knowledge Distillation for Surgical Phase Recognition},
  author={Zhang, Jinglu and Barbarisi, Santiago and Kadkhodamohammadi, Abdolrahim and Stoyanov, Danail and Luengo, Imanol},
  journal={International Journal of Computer Assisted Radiology and Surgery},
  year={2024},
  note={arXiv:2306.08961}
}

@inproceedings{guan2024label,
  title={Label-Guided Teacher for Surgical Phase Recognition via Knowledge Distillation},
  author={Guan, Jiale and Zou, Xiaoyang and Tao, Rong and Zheng, Guoyan},
  booktitle={Medical Image Computing and Computer-Assisted Intervention -- MICCAI 2024},
  pages={349--358},
  year={2024},
  publisher={Springer},
  doi={10.1007/978-3-031-72089-5_33}
}

@article{qwenteam2025qwen3vl,
  author  = {{Qwen Team}},
  title   = {Qwen3-VL Technical Report},
  journal = {arXiv preprint arXiv:2511.21631},
  year    = {2025}
}

@article{liu2023visual,
  title={Visual instruction tuning},
  author={Liu, Haotian and Li, Chunyuan and Wu, Qingyang and Lee, Yong Jae},
  journal={Advances in neural information processing systems},
  volume={36},
  pages={34892--34916},
  year={2023}
}

@article{abouelenin2025phi4multimodal,
  title={Phi-4-Mini Technical Report: Compact yet Powerful Multimodal Language Models via Mixture-of-LoRAs},
  author={Abouelenin, Abdelrahman and others},
  journal={arXiv preprint arXiv:2503.01743},
  year={2025}
}

@misc{mistral2025small31,
  author={{Mistral AI}},
  title={Mistral Small 3.1},
  year={2025},
  howpublished={\url{https://mistral.ai/news/mistral-small-3-1}}
}

@article{zhu2025internvl3,
  title={InternVL3: Exploring Advanced Training and Test-Time Recipes for Open-Source Multimodal Models},
  author={Zhu, Jinguo and Wang, Weiyun and Chen, Zhe and Liu, Zhaoyang and Ye, Shenglong and others},
  journal={arXiv preprint arXiv:2504.10479},
  year={2025}
}

@article{yao2024minicpmv,
  title={MiniCPM-V: A GPT-4V Level MLLM on Your Phone},
  author={Yao, Yuan and others},
  journal={arXiv preprint arXiv:2408.01800},
  year={2024}
}

@misc{openbmb2024minicpmo26,
  author={OpenBMB},
  title={MiniCPM-o 2.6},
  year={2024},
  howpublished={\url{https://huggingface.co/openbmb/MiniCPM-o-2_6}}
}

@article{gemmateam2025gemma3,
  title={Gemma 3 Technical Report},
  author={{Gemma Team}},
  journal={arXiv preprint arXiv:2503.19786},
  year={2025}
}

@article{peng2025skyworkr1v,
  title={Skywork R1V: Pioneering Multimodal Reasoning with Chain-of-Thought},
  author={Peng, Y. and others},
  journal={arXiv preprint arXiv:2504.05599},
  year={2025}
}

@misc{meta2025llama4,
  author={{Meta AI}},
  title={The Llama 4 Herd: The Beginning of a New Era of Natively Multimodal AI Innovation},
  year={2025},
  howpublished={\url{https://ai.meta.com/blog/llama-4-multimodal-intelligence/}}
}

@article{qwenteam2025qwen25vl,
  title={Qwen2.5-VL Technical Report},
  author={{Qwen Team}},
  journal={arXiv preprint arXiv:2502.13923},
  year={2025}
}

@article{hurst2024gpt4o,
  title={GPT-4o System Card},
  author={Hurst, Aaron and others},
  journal={arXiv preprint arXiv:2410.21276},
  year={2024}
}

@article{geminiteam2025gemini,
  title={Gemini 2.0: Technical Report},
  author={{Gemini Team}},
  journal={arXiv preprint arXiv:2501.06252},
  year={2025}
}

\end{document}